%% file: main.tex
\documentclass[letterpaper, 10 pt, conference]{ieeeconf}

\usepackage{etoolbox}
\makeatletter
\patchcmd{\@makecaption}
  {\scshape}
  {}
  {}
  {}
\makeatletter
\patchcmd{\@makecaption}
  {\\}
  {.\ }
  {}
  {}
\makeatother

\IEEEoverridecommandlockouts                              % This command is only needed if you want to use the \thanks command
\usepackage{times}

\usepackage{multicol}
\usepackage[bookmarks=true]{hyperref}
\usepackage{graphicx}
\usepackage{tabularx}
\usepackage{amssymb}
\usepackage{amsmath}
\usepackage{fancyhdr}
\usepackage{tikz}
\usepackage{bm}
\usepackage{listings}
\usepackage{subfigure}
\usepackage{xcolor}
\usepackage{scalerel}
\usepackage[colorinlistoftodos]{todonotes}
\usepackage{algorithm}
\usepackage[noend]{algpseudocode}
\usepackage{algorithmicx}
\usepackage{float}
\usepackage{tikz}
\usepackage{graphics} 
\usetikzlibrary{positioning}
\usetikzlibrary{shapes,arrows}
\usepackage{dcolumn}
\usepackage{multirow}
\usepackage{makecell}
\usepackage[font=small,skip=2pt]{caption}
\newcommand{\vectg}[1]{\boldsymbol{#1}}

\newcommand{\bx}{\mathbf{x}}

\newcommand{\bu}{\mathbf{u}}

\newcommand{\bp}{\mathbf{p}}

\newcommand{\bz}{\mathbf{z}}
\newcommand{\bX}{\boldsymbol{\tau}}
\newcommand{\bU}{\boldsymbol{\xi}}

\newcommand{\bZ}{\mathbf{Z}}
\newcommand{\bnu}{\boldsymbol{\nu}}
\newcommand{\bbeta}{\boldsymbol{\eta}}

\newcommand{\bpi}{\boldsymbol{\pi}}
\newcommand{\bmu}{\boldsymbol{\mu}}
\newcommand{\bzeta}{\boldsymbol{\zeta}}
\newcommand{\bSigma}{\boldsymbol{\Sigma}}

\def\NAT@parse{\typeout{IEEEtran error: Attempt to use fake Natbib command 
which is provided to fool Hyperref.}}

\allowdisplaybreaks
\renewcommand{\baselinestretch}{0.98}
\begin{document}

\onecolumn
\newpage
\thispagestyle{empty}  % No page number
\begin{center}
    \vspace*{\fill}  % Vertically center the text
    \textcopyright 2025 IEEE. Personal use of this material is permitted. Permission from IEEE must be obtained for all\\
    other uses, in any current or future media, including reprinting/republishing this material for advertising\\
    or promotional purposes, creating new collective works, for resale or redistribution to servers or lists,\\
    or reuse of any copyrighted component of this work in other works.
    \vspace*{\fill}  % Vertically center the text
\end{center}

\newpage 
\twocolumn% Start the next page for your paper

% TITLE / AUTHORS
\title{\LARGE \bf A Multimodal Stochastic Planning Approach for Navigation and Multi-Robot Coordination}
%alternate titles: MultiModal Stochastic Trajectory Optimization for Multi-Robot Planning
% Multimodal Sampling Based Planner for Multi-Agent Motion Planning in Trap Environments
% Real-Time Multimodal Planning for Navigation and Multi-Robot Collision Avoidance
% Real-Time Multimodal Planning for Navigation and Multi-Robot Coordination
% Multimodal Receding Horizon Control for Multi-Robot Planning
% Multimodal Policy Optimization for Multi-Robot Planning
\author{Mark Gonzales, Ethan Oh, Joseph Moore
\thanks{Johns Hopkins University \newline \hspace*{1.6em}
{\tt\small \{MGonza60, EOh18, JLMoore\}@jhu.edu}} }
% \newline \hspace*{0.8em} $^{2}$Johns Hopkins University Applied Physics Lab \newline \hspace*{1.6em} {\tt\small \{Joseph.Moore\}@jhuapl.edu}}}
\maketitle

% ABSTRACT
\begin{abstract}

%Local minima present a significant challenge in trajectory optimization, particularly for finite-horizon planners constrained by environmental factors, robot dynamics, and interactions among multiple agents. 

In this paper, we present a receding-horizon, sampling-based planner capable of reasoning over multimodal policy distributions. By using the cross-entropy method to optimize a multimodal policy under a common cost function, our approach increases robustness against local minima and promotes effective exploration of the solution space. We show that our approach naturally extends to multi-robot collision-free planning, enables agents to share diverse candidate policies to avoid deadlocks, and allows teams to minimize a global objective without incurring the computational complexity of centralized optimization. Numerical simulations demonstrate that employing multiple modes significantly improves success rates in trap environments and in multi-robot collision avoidance. Hardware experiments further validate the approach's real-time feasibility and practical performance.

\end{abstract}
\IEEEpeerreviewmaketitle

% SECTIONS
\input{introduction}

\input{related_work}

\input{background}
\input{approach}

\input{sim_experiments}

\input{hardware_experiments}
\input{discussion}

% BIBLIOGRAPHY
\bibliographystyle{IEEEtran}
% \clearpage
\bibliography{references}

% \clearpage
% \input{appendix}

\end{document}

%% file: introduction.tex
\section{Introduction}
Local minima pose a fundamental challenge for finite-horizon, gradient-based planning approaches. In multi-robot scenarios, local minima can arise not only from the environment but also from dynamic factors, such as the changing trajectories of teammates, which may inadvertently block or cut off routes that would otherwise be viable. These pitfalls often cause robots to become stuck, find suboptimal solutions, or fail to coordinate effectively in complex environments.

Sampling-based planners, such as Model Predictive Path Integral (MPPI)~\cite{Williams2016} and Cross-Entropy Method (CEM)~\cite{Rubinstein1999, Kobilarov2012}, attempt to improve the trajectory cost by stochastically sampling and evaluating trajectories in the cost landscape. In practice, these methods utilize hyperparameters, such as sampling variance, number of samples, and the horizon length, to adapt the exploration to the environment. However, both MPPI and CEM typically sample trajectories around the prior best policy, leading to a concentration of samples in a narrow region of the solution space. This localized search impedes the planner’s ability to effectively navigate around traps or escape from local minima once they occur, especially in environments with challenging topology. As a result, the planner can become stuck in suboptimal regions, regardless of the variance or adaptation strategy.

In multi-robot systems, the difficulty is exacerbated by the need for robots to coordinate planned trajectories. Centralized control approaches~\cite{Sandip2012, Wu2021, Beyoglu2022, Liu2017} can, in principle, achieve globally optimal coordination; however, they suffer from scalability issues and high computational costs as the team size increases. Distributed methods, while scalable, often require robots to individually select their optimal trajectory, subsequently negotiating with teammates to reach a feasible consensus. When each robot contributes only a single candidate trajectory, the team risks deadlock or persistent local minima, as a lack of trajectory diversity reduces the likelihood of discovering collision-free, cooperative maneuvers, especially when teammates dynamically update their plans or block each other’s routes in real-time.

To overcome these limitations, we introduce a multimodal sampling and clustering framework that maintains multiple policy candidates for each robot, thereby increasing diversity and robustness against local minima in both environmental and collaborative planning contexts.
\begin{figure}[t]
    \centering
     \includegraphics[width=\columnwidth]{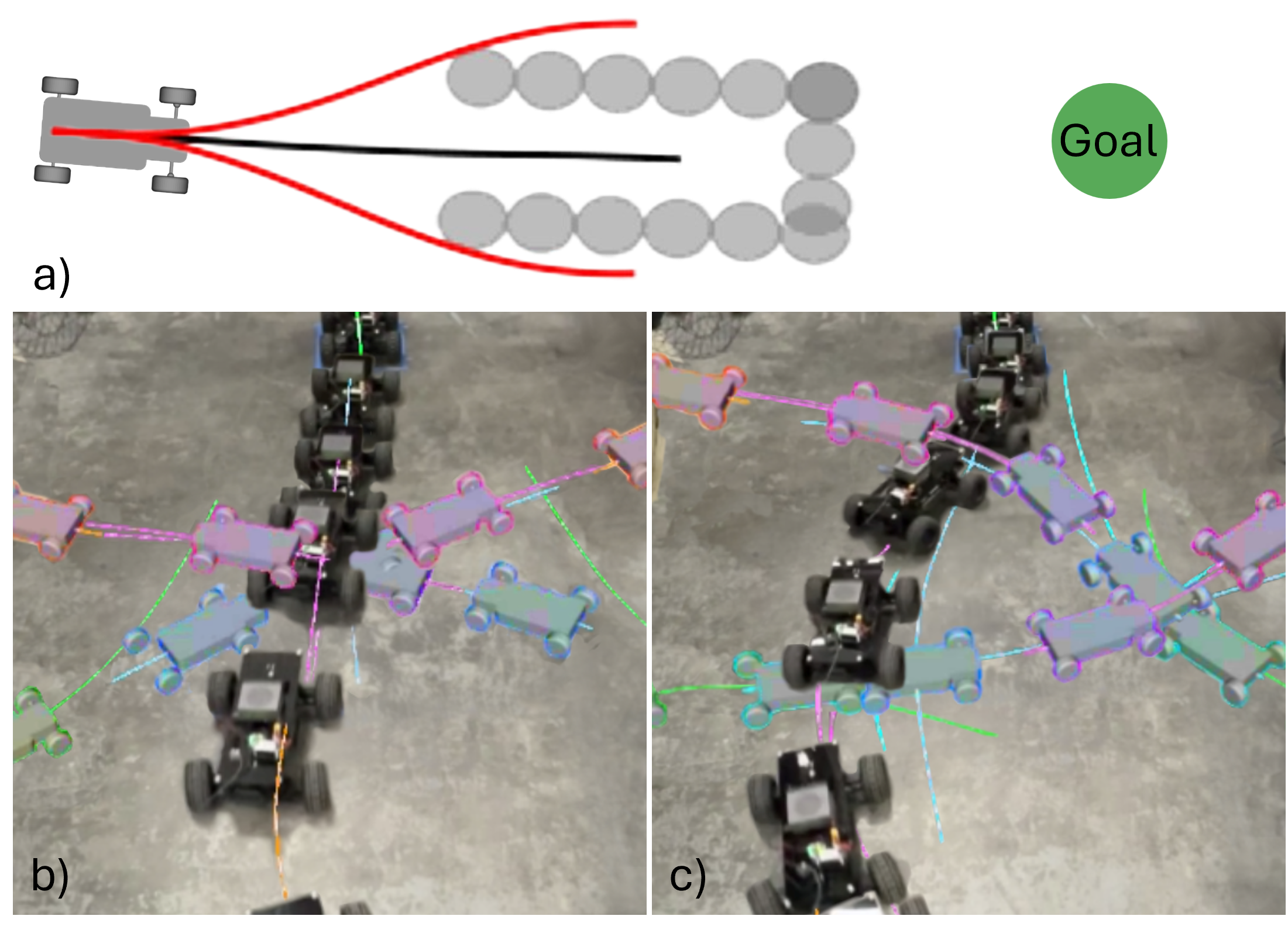}
    \caption{a) A visualization of a multimodal policy with three modes. b) Timelapse of Rally car failing to avoid other virtual robots using one mode. c) Timelapse of Rally car successfully avoiding other virtual robots using two modes.}
    \label{fig:Wedge}
\end{figure}
Our contributions are:
\begin{itemize}
    \item A cross-entropy planning approach capable of preserving multiple policy modes for increased planning robustness.
    \item A multi-robot coordination framework that enables reasoning about sets of candidate policies to avoid local minima and deadlocks more reliably.
\end{itemize}

% \begin{figure}[t]
%     \centering
%     \includegraphics[width=\columnwidth]{images/5Wedge.png}
%     \caption{A five-agent team in a wedge formation in an obstacle field.}
%     \label{fig:Wedge}
%     \vspace{-15pt}
% \end{figure}

% \begin{figure}[t]
%     \centering
%     \includegraphics[width=\columnwidth]{images/MainImage2.png}
%     \caption{Top: A five-agent team in a wedge formation in an obstacle field. Bottom: Two Fixed Wing Avoiding a Head on Collision}
%     \label{fig:Wedge}
% \end{figure}

%% file: related_work.tex
\section{Related Work}
This section reviews recent advances in multimodal policy optimization and multi-robot planning, which jointly motivate the approach proposed in this work.

\subsection{Multi-Modal Policy Optimization}
Numerous approaches have been proposed to address the multimodality inherent in trajectory optimization. In reinforcement learning, multi-policy and ensemble strategies are explored, such as maintaining a buffer of diverse candidate policies while pruning redundant, low-performing policies to improve exploration and diversity~\cite{Pan2021}, or leveraging latent representations to discover richer policy landscapes~\cite{Huang2023}.

Diffusion models have recently gained traction for their ability to generate diverse trajectory distributions and systematically explore the trajectory space, thereby avoiding local minima, particularly in cases of cul-de-sacs~\cite{Knuth2024}. Alternative methods handle multimodal path planning by explicitly sampling both trajectories and cost functions~\cite{Carvalho2025}.

% Other approaches sample trajectories and cost functions for explicitly handling multimodal path planning~\cite{Carvalho2025}.

In settings with uncertain future behaviors of dynamic agents, belief estimation and event likelihoods are exploited to optimize over multiple candidate policies, often by generating samples under differing cost functions that reflect possible environment predictions~\cite{Cunningham2015, Zhou2018}. Parallel, sample-based policy optimization has also been implemented, where each instance targets a distinct goal or environmental hypothesis~\cite{Zhang2024, Zhou2025, Trevisan2024}.

Other random sampling-based approaches attempt to map the cost landscape by identifying and retaining promising modes, or, in some cases, facilitating more rapid collapse toward a single optimal solution~\cite{Osa2020, Honda2024, Okada20}. Other techniques, including the application of variational autoencoders, have been explored to capture the manifold of local minima in the planning landscape and to guide sampling in latent spaces~\cite{Osa2022}. Other approaches sample global plans by partitioning the environment to find all the homotopy classes~\cite{Rosmann_2016}
% These have been used for global path planning.

\subsection{Real-Time Planning for Multi-Robot Coordination}
%Control of multi-robot systems in environments with obstacles has been addressed through various approaches. Consensus-based methods enable coordination by sharing local state or intention information among robots, as seen in~\cite {Mylvaganam2017, Cappello2021}, which utilize this combined information to find Nash Equilibrium solutions. Other approaches that share information mutually determine no-go zones or assign a priority to a robot~\cite{Vrba2007}. Collision avoidance in multi-robot systems has also been tackled using behavior-based and decentralized control strategies~\cite{Balch1998, Emile2020, Saber2003, Sa2023, Chang2003}. However, such methods typically require access to the goals or intended actions of all robots, and are generally limited to systems with simple or kinematic dynamics.
Over the years, many approaches for multi-robot coordination and control have been proposed \cite{verma2021multi}. Recently, nonlinear model predictive control (NMPC) has become a popular framework for trajectory optimization and planning in multi-robot systems~\cite{Roldao2014, Xiao2016}. Extensions include NMPC combined with conflict-based search to efficiently identify collision-free solutions~\cite{Tajbakhsh2024}, as well as priority-based schemes where planning times are offset to allow sequential path negotiation~\cite{Tordesillas2020}. Distributed NMPC approaches have also emerged, either by sampling and sharing predicted states from other robots' planned trajectories~\cite{Satir2023}; by integrating on-demand collision avoidance mechanisms within the NMPC optimization loop~\cite{Luis2020}; or by prioritizing collision avoidance based on more urgent predicted collisions~\cite{Lindqvist2021}. \cite{Madabushi2025} presents a distributed tube-based NMPC approach for aerial vehicle swarms, while \cite{Gonzales2025} combines a stochastic NMPC approach, PAC-NMPC \cite{Polevoy2023}, with an objective function inspired by gyroscopic obstacle avoidance to address challenges with deadlock in distributed NMPC. 

In this paper, we present an approach for real-time stochastic multimodal policy optimization to improve both single-robot navigation and multi-robot collision avoidance. To our knowledge, this is the first approach to use multimodal stochastic NMPC for coordinated multi-robot maneuvers.

%% file: background.tex
\section{Problem Formulation}
\label{sec:prob_form}
Here, we present a formulation for the single and multi-robot stochastic optimization problems.
\subsection{Single Robot Planning}
Consider a robot whose stochastic dynamics are defined by the probability density function $p(\bx_{t+1}|\bx_{t}, \bu_{t})$, where $\bx_{t}\in\mathbb{R}^{N_x}$ is a vector of state values and $\bu_{t}\in\mathbb{R}^{N_u}$ is a vector of control inputs. Given a policy $\bu_t=\bpi(\bx_t,\bU)$ with policy parameters $\bU$, we can write a trajectory distribution
\begin{align}
p(\bX|\bU)~=~p(\bx_0)\prod^T_{t=0}p(\bx_{t+1}|\bx_{t},\bu_t)
\label{eq:stochastic_dynamics}
\end{align} 
where $\bX$ is a discrete-time trajectory sequence $\{\bx_0,\bu_0, \bx_1,\bu_1 ..., \bu_{N_T}, \bx_{N_T+1} \}$ and $N_T$ is the number of timesteps. Our goal is to minimize the cost $J(\bX) = \sum_{t=0}^{N_T-1} q(\bx_t, \bu_t) + q_f(\bx_{N_T})$ such that constraint $C(\bX)\le 0$. Here, $q(\bx_t, \bu_t)$ is the running cost at time $t$ and $q_f(\bx_{N_T})$ is the cost at the final time. 

To ensure collision avoidance and adhere to state bounds, we formulate the constraints as
% if any of these are true, we are violating constraints.
\begin{align}
    g_b(\bx_t) &= (\bx_t - \bx_l) < 0 \lor (\bx_u - \bx_t) < 0 \\
    g_{o}(\bx_t) &= \{ \text{dist}(\bx_t, \bp^{o^m}) - r < 0 \} \ \forall \ m \\
    % robot-to-robot constraint function g_A could be added here
    c(\bx_t) &= g_b(\bx_t) \le 0 \lor g_{o}(\bx_t) \le 0 \label{eq:constraint} \\
    C(\bX) &= c(\bx_0) \lor c(\bx_1) \cdots \lor c(\bx_{N_T}) 
\end{align}
where $\bx_{l}$ and $\bx_{u}$ are the lower and upper state bounds respectively, $\bp^{o^m}$ is the $m^{\textrm{th}}$ obstacle position, and $r$ is the collision radius. We then seek to enforce the probability of constraint violation such that $\mathbb{E} [C(\bX)] \le P$ where $P\in[0,1]$.

%In this paper, we formulate all constraints as chance constraints of the form
%\begin{align}
%    C_A^{n, k}(\bX^i) &= \mathbb{I}\left( g_A^{n, k}(\bX^i) \geq P \right)
%\end{align}

\subsection{Multi-Robot Planning}
For the multi-robot planning problem, we include additional inter-robot constraints. Let $\bx_t^i$ be the state of the $i^{\textrm{th}}$ robot at time $t$. We then define $g_a$ as a constraint to prevent collisions between dynamic agents and given as:
\begin{align}
g_a(\bx_t^i) &= \{dist(\bx_t^i, \bx_t^n) - L < 0\} \ \forall \ n: i \neq n,%\\
\end{align}
where $dist()$ is the euclidean distance between the robots, $L$ is the collision radius, and $\bx_t^n$ is $n^{\textrm{th}}$ robot on the team. We can then modify Eq. \ref{eq:constraint} for the $i^{\textrm{th}}$ robot as follows:
\begin{align}
    c(\bx_t^i) &= g_b(\bx_t^i) \le 0 \lor g_{o}(\bx_t^i) \le 0 \lor g_{a}(\bx_t^i) \le 0.
\end{align}
\section{Background}
Here we review the Cross-Entropy Method (CE) \cite{Rubinstein1999}, a sample-based stochastic optimization algorithm, and its application to trajectory optimization under complex cost and constraint landscapes~\cite{Kobilarov2012}.
\subsection{Cross-Entropy Optimization}
\label{sec:CEOpt}
Consider a cost function $J(\bz)$, where parameter $\bz$ is an instance of the random variable $\bZ$ described by probability density function $p(\bz,\bar{\bnu})$ with hyperparameters $\bar{\bnu}$. CE attempts to estimate the rare event probability
\begin{align}
\ell = \mathbb{P}_{\bar{\bnu}}(J(\bZ)\le \gamma) = \mathbb{E}_{\bar{\bnu}}[\mathbb{I}_{J(\bZ)\le\gamma}],   
\end{align} where $\gamma$ is a small positive constant and $\mathbb{I}$ is the indicator function. Instead of sampling using true hyperparameters $\bar{\bnu}$, we can use importance sampling to write
\begin{align}
\ell = \mathbb{E}_{\bnu} \left[ \mathbb{I}_{J(\bZ)\le\gamma} \frac{p(\bZ,\bar{\bnu})}{p(\bZ,\bnu)} \right].
\end{align}
The optimal distribution, with hyperparameters $\bnu^*$, that minimizes the estimator variance, is given as  $p(\bz,\bnu^*) = \frac{\mathbb{I}_{J(\bZ)\le\gamma}p(\bz,\bar{\bnu})}{\ell}$. Since $\ell$ is unknown, CE seeks a tractable proposal distribution $p(\bz,\bnu)$, that approximates $p(\bz,\bnu^*)$ by minimizing the Kullback-Leiber (KL) divergence
\begin{align}
\bnu^* &= \arg\min_{\bnu} KL(p(\bz,\bnu^*), p(\bz,\bnu)).
%&= \arg\max_v \mathbb{E}_{q^*}[\log p(Z,v)] \nonumber\\ 
\end{align}
By applying the definition of the KL divergence and substituting in for $p(\bz,\bnu^*)$, we have
\begin{align}
\bnu^*= \arg\max_{\bnu} \mathbb{E}_{\bar{\bnu}}\left[ \mathbb{I}_{J(\bZ) \leq \gamma} \log p(\bZ,\bnu)\right],
\label{eq:importance}
\end{align}
which can be approximated via sampling by
\begin{align}
\hat{\bnu}^* = \arg\max_v \frac{1}{N} \sum_{i=1}^N \mathbb{I}_{J(\bZ_i) \leq \gamma} \log p(\bZ_i, \bnu).
\end{align}

The multi-level CEM algorithm adaptively updates the threshold $\gamma$ using elite samples as follows:
\begin{enumerate}
    \item Sample $\bZ_1, \ldots, \bZ_N \sim p(\bz, \hat{\bnu}_{j-1})$ and set $\gamma$ to the $\rho$-quantile of $J(\bZ)$.
    \item Update parameters such that %\hat{\bnu}_j$
    \begin{align}
        \hat{\bnu}_j = \arg\max_{\bnu} \frac{1}{N} \sum_{i=1}^N \mathbb{I}_{J(\bZ_i) \leq \gamma} \log p(\bZ_i, \bnu).
            \label{eq:CEalgorithm}
    \end{align}
\end{enumerate}

% The CEM iteratively samples $M$ control trajectories from a multivariate Gaussian distribution, for each control at each timestep, and then uses the dynamics model to get an associated state trajectory for each sample. Then it evaluates the cost and constraints violations for each trajectory, and based on the costs for each trajectory, we select the top $N\%$ of the trajectories as the elite control trajectory samples. These elite control trajectory samples are used to update the parameters of the multivariate Gaussian, by taking the mean and variance of the elite distribution; this process is then repeated, focusing future exploration around more promising regions of the policy space.

\subsection{Cross-Entropy Trajectory Optimization}
\label{sec:CETrajOpt}
As described in \cite{Kobilarov2012}, CE can be used for stochastic trajectory optimization. We present this approach in the context of the problem formulation in Section \ref{sec:prob_form}.

Consider the stochastic dynamics $p(\bx_{t+1}|\bx_{t}, \bu_{t})$ as defined in Section \ref{sec:prob_form}. Let us parameterize an open-loop control policy $\bu_t =\bpi(\bU) =\bzeta_t$, where $\bU ~= ~\begin{bmatrix} \bzeta_0^T & \bzeta_1^T& ...&  \bzeta_{N_T}^T\end{bmatrix}^T$ and $\bzeta_t \in \mathbb{R}^{N_u}$. Let us also parameterize a distribution $p(\bU|\bnu)$ as a multivariate Gaussian over the discrete-time control sequence so that $ \bU \sim \mathcal{N}\left(\bU | \bmu, \bSigma \right)$. For computational tractability, we will also assume a diagonal covariance matrix $\bSigma$ so the distribution parameters are given as $\bnu \triangleq \begin{bmatrix} \bmu^T, diag(\bSigma)^T\end{bmatrix}^T$ where $ diag(\bSigma) = \begin{bmatrix} \bbeta_0^T & \bbeta_1^T& ...&  \bbeta_{N_T}^T\end{bmatrix}^T$ and $\bbeta_t \in \mathbb{R}^{N_u}$. We can then write a joint distribution
$p(\bX, \bU|\bnu)=p(\bX|\bU)p(\bU|\bnu)$, where $p(\bX|\bU)$ is given by Equation \ref{eq:stochastic_dynamics}.

Given a cost function $J(\bX)$, where $\bX\sim p(\bX, \bU|\bnu)$, we can use the CE approach described in  \ref{sec:CEOpt} to optimize the distribution by letting $\bX=\bz$. Algorithm~\ref{alg:opti_loop} outlines the adaptation of CE for sample-based trajectory optimization.

%Thus, a control trajectory with an $N_u$-dimensional control input and $N_T$ timesteps, the surrogate distribution is parameterized as a $N_u N_T$-dimensional multivariate Gaussian where  $\bmu \in \mathbb{R}^{N_u N_T}$ and $diag(\bSigma) \in \mathbb{R}^{N_u N_T}$.

%To use CE for trajectory optimization, we parameterize the proposal distribution $p(z,v)$ as a multivariate Gaussian over the control sequence $(\mu, \Sigma)$ for a fixed planning horizon. At each iteration, the algorithm samples control trajectories and propagates the control trajectories using a dynamics model and evaluates the trajectories based on cost and constraint functions. 
\begin{algorithm}[H]
\caption{CE for Trajectory Optimization}
\label{alg:opti_loop}
% optimization loop
\begin{algorithmic}[1]
    \State Initialize distribution parameters $\bnu$
    \While{termination condition not met}
        \For{$j = 1$ to $M$}
            \State Sample control sequence  $ \bU \sim \mathcal{N}\left(\bU | \bnu \right)$
            \State Simulate trajectory $\bX\sim p(\bX, \bU|\bnu)$
            \State Evaluate cost $J_j = J(\tau)$
            \State Evaluate constraints $C_j = C(\tau)$
        \EndFor
        \State Select elite set: $\mathcal{E} = \textrm{top } \rho\% \textrm{ samples}$
        \State Update $(\bnu)$ using elite $\mathcal{E}$
    \EndWhile
    \State \textbf{Output:} Best trajectory $\bX^*$
\end{algorithmic}
\end{algorithm}

%% file: approach.tex
\section{Multimodal Cross-Entropy Planning}
In this section, we present a modified CE algorithm for real-time planning in environments that give rise to multimodal policy distributions. 
\subsection{Multimodal Policy Parameterization}
To parameterize our multimodal policy, $\bpi(\bU)$, we choose a Gaussian Mixture Model (GMM) such that
\begin{align}
p(\bU|\bnu) =  \sum_{k=1}^K \phi_k \mathcal{N}(\bU|\bmu_k,\bSigma_k)
\end{align}
where $\bnu \triangleq \begin{bmatrix} \bmu_1^T, diag(\bSigma_1)^T, \ldots \bmu_K^T, diag(\bSigma_K)^T\end{bmatrix}^T$, $\phi_k$ are scalar weights, and $K$ is the number of modes. Similar to Section \ref{sec:CETrajOpt}, each mode is represented by a $N_u N_T$-dimensional multivariate Gaussian, where  $\bmu_k \in \mathbb{R}^{N_u N_T}$ and $diag(\bSigma_k) \in \mathbb{R}^{N_u N_T}$.

Given $p(\bX|\bU)$, sampling trajectory sequences $\bX$ using this multimodal policy distribution is then straightforward, since $p(\bX,\bU|\bnu) = p(\bX|\bU)p(\bU|\bnu)$. However, if we wish to preserve multimodal policies during planning, Algorithm \ref{alg:opti_loop} often cannot be directly applied.

\subsection{Challenges with Multimodal CE Planning}
One challenge associated with planning multimodal policies using CE trajectory optimization is that the standard CE algorithm can lead to mode collapse, even in a single step of the algorithm, when two feasible policies result in significantly different costs. Consider the case where $D$ represents a finite set of feasible trajectory samples, and where $D = D1 \cup D2$ and $D1 \cap D2 = \emptyset $. Let $D1$ be contained in mode $M1$, such that $D1\subset M1$. Let $D2$ be contained in $M2$, such that $D2\subset M2$. We also assume $M1 \cap M2 = \emptyset $. Now, let the set of costs for $D1$ be denoted as $J(D1)$ and the set of costs for $D2$ be denoted as $J(D2)$. If, for a given environment, $\sup J(D1) < \inf J(D2)$ and $|D1|> |S_{\mathcal{E}}|$, where $S_{\mathcal{E}}$ is the set of elite samples, it follows from Algorithm \ref{alg:opti_loop}, that $S_{\mathcal{E}} \subset D1$ and the resulting policy will collapse into $M1$.

% Consider the case where $D$ represents a finite set of feasible trajectory samples, with $D = D_1 \cup D_2$ and $D_1 \cap D_2 = \emptyset$. Let $D_1$ be contained in mode $M_1$, such that $D_1 \subset M_1$, and let $D_2 \subset M_2$, where $M_1 \cap M_2 = \emptyset$. Denote by $J(D_1)$ and $J(D_2)$ the sets of cost values associated with $D_1$ and $D_2$, respectively. If, for a given environment, $\sup J(D_1) < \inf J(D_2)$ and $|D_1| > |S_{\mathcal{E}}|$, where $S_{\mathcal{E}}$ is the set of elite samples, then by Algorithm~\ref{alg:opti_loop}, it follows that $S_{\mathcal{E}} \subset D_1$, and the resulting policy will collapse into $M_1$.

For this reason, we modify the CE algorithm to preserve higher-cost modes, as these modes may become lower-cost modes under finite-horizon planning as the horizon recedes.  

%First, since CE trajectory optimization is restricted to a finite number of samples, it is possible that algorithm will be unable to fully explore the multimodal nature of the environment. 

\subsection{Trajectory Clustering and Feasibility Sampling}
Consider the CE parameter update equation \ref{eq:CEalgorithm}, now modified for the CE trajectory optimization:
  \begin{align}
        \hat{\bnu}_j = \arg\max_{\bnu} \frac{1}{N} \sum_{i=1}^N \mathbb{I}_{J(\bX_i) \leq \gamma} \log p(\bX_i| \bnu).
        \label{eq:CETrajOptUpdate}
    \end{align}
While closed-form solutions exist for the maximum-likelihood estimation of unimodal normal distribution parameters, such solutions do not exist for a GMM, and heuristic solutions have been proposed \cite{bishop2006pattern}. Therefore, to preserve the multimodal nature of our policy distribution, we propose sampling from the set of feasible (constraint-free) trajectories and applying Equation \ref{eq:CETrajOptUpdate} to distinct modes. 
\subsubsection{Feasibility Sampling}
For our receding-horizon navigation and obstacle avoidance tasks, we are primarily concerned with multimodal policies that yield constraint-free trajectories. To this end, we sample from our policy and only preserve constraint-free samples. In the case where no constraint-free trajectories exist, the constraints are relaxed, and the total number of violations is incorporated as an additive penalty to the cost. All trajectories are then used for clustering.
\subsubsection{Trajectory Clustering}
We identify distinct solution modes in these feasible samples via the K-means clustering algorithm~\cite{MacQueen1967}. K-means is an unsupervised learning method that partitions data into $K$ clusters by iteratively assigning each sample to its nearest centroid (chosen randomly from the data at initialization), then updating centroids to be the mean of each cluster until convergence. In particular, we cluster trajectories based on state sequences, which promotes the discovery of distinct homotopy classes or task-relevant policy modes. 
%In our framework, each constraint-free trajectory is represented as a flattened state sequence and grouped according to similarity in state space. 
Let ${\boldsymbol{\chi}}_n = \begin{bmatrix}\bx_{n,0}^T  &\bx_{n,1}^T ...&\bx_{n,N_T+1}^T\end{bmatrix}^T$, where $\bx_{n,k}$ represents the $k^{\textrm{th}}$ time sample of the state $n^{\textrm{th}}$ feasible trajectory sample. K-means clustering seeks $K$ cluster centroids $\{\mathbf{y}_1, \dotsc, \mathbf{y}_K\}$ by iteratively performing ${z_n~= \arg\min_{k} \left\| {\boldsymbol{\chi}}_n - \mathbf{y}_k \right\|_2^2 \hspace{5mm} \forall n}$ and $\mathbf{y}_k = \frac{1}{|S_k|} \sum_{n \in S_k} {\boldsymbol{\chi}}_n$ where $S_k = \left\{ n : z_n = k \right\}$ is the set of trajectories assigned to cluster $k$, and $\|\cdot\|_2$ denotes the Euclidean norm.
\subsubsection{Multimodal Policy Update}
Given the policy modes, we apply the CE update step to each cluster. Within each mode, we select the top $\rho$-quantile elite samples by cost and use MLE to fit new Gaussian parameters to their control policies. Equation \ref{eq:CETrajOptUpdate} becomes
\begin{align}
    % J_k(\bx) &= \\
    \hat{\bnu}_j &= \arg\max_{\bnu}\frac{1}{|S_k|}\sum_{i\in S_k}\mathbb{I}_{J_k(\bX_i) \leq \gamma} \log p(\bX_i| \bnu).
\end{align}

% \begin{enumerate}
%     \item Sample $Z_1, \ldots, Z_N \sim p(z, \hat{v}_{j-1})$ and set $\gamma$ to the $\rho$-quantile of $J(Z)$.
%     \item Update parameters: 
%     \begin{align}
%         \hat{v}_j = \arg\max_v \frac{1}{N} \sum_{i=1}^N \mathbb{I}_{J(Z_i) \leq \gamma} \log p(Z_i, v).
%     \end{align}
% \end{enumerate}
    
%The updated policies are then used as the candidate distributions for sampling in the next planning iteration. This cycle repeats until a termination condition is met, at which point the policy with the lowest average cost is selected for execution as seen in Algorithm~\ref{alg:multi_modal}
The complete algorithm for Multimodal CE Trajectory Optimization can be seen in Algorithm~\ref{alg:multi_modal}.
\vspace{1em}
\begin{algorithm}
\caption{Multimodal Cross Entropy}
\label{alg:multi_modal}
\begin{algorithmic}[1]
    \State \textbf{Inputs:} $\hat{\nu}_0$
    \While {termination conditions not met}
        \For{$j = 1$ to $M$}
            \State Sample control sequence $\bU \sim p(\bU|\bnu)$
            \State Simulate trajectory $\bX \sim p(\bX|\bU)$, 
            \State Evaluate cost $J_j = J(\bX)$
            \State Evaluate constraints $C_j = C(\bX)$
        \EndFor
        \State Filter feasible set $\mathcal{T}_{\textrm{free}} = \{ \bX_j : C_j = 0 \}$
        \State Cluster feasible trajectories using K-means$(\mathcal{T}_{\textrm{free}}, K)$
        \For{each cluster $k$}
            % \State Select elite set: $\mathcal{S}_{k,\textrm{elite}} = \mathrm{Top}_{\rho\%}(\mathcal{S}_k)$
            \State Select elite set: $\mathcal{E}_{k} = \mathrm{Top}_{\rho\%}(\mathcal{S}_k)$
            
            \State Update $(\bmu_k, \bSigma_k)$ from $\mathcal{E}_{k}$
        \EndFor
    \EndWhile
\end{algorithmic}
\end{algorithm}
\subsection{Warm Starts for Secondary Modes}
To accelerate convergence and improve sample efficiency in iterative replanning, we employ warm-start initialization strategies for both primary and secondary policy modes. For the primary mode, we warm start the new policy by shifting the prior policy forward by the number of executed steps.
%, 
%$n_{\textrm{exec}}$:
%\begin{align}
 %   \bu_t \gets \bu_{t + n_{\textrm{exec}}}, \ t=0, \cdots, N_T - n_\textrm{exec} - 1\\
%    \bu_t \gets 0.0, \ t= N_T - n_\textrm{exec}, \cdots, N_T - 1
%\end{align}
For each secondary mode, naively shifting controls can result in poor initialization, since the robot is not following those control inputs. To preserve these modes, a time-varying linear quadratic regulator (TVLQR) policy on the prior nominal trajectory is used to calculate gains $\boldsymbol{\kappa}_t$ for each time step. Given the current state $\bx_0$, and mode $k$, the warm-started control at step $t$ is $\bu_t^* = \bu_t^k + \boldsymbol{\kappa}_t^k(\bx_t - \bx_t^k)$. %, \qquad t = 0,\ldots,N_T-1$.

\begin{algorithm}
\caption{Secondary Mode Warm Start}
\label{alg:warm_start}
    \begin{algorithmic}[1]
        \State Inputs: $\mathbf{\bX}^k, \bx_0$; 
        \State $\bX^k = \{\bx_0^k, \bu_0^k,\bx_1^k, \bu_1^k, \cdots, \bu_{N_T}^k,\bx_{N_T+1}^k\}$ 
        \State $\boldsymbol{\kappa} = \text{TVLQR}(\bX^k)$;
        \For{$t = 0, \cdots, N_t$}
            \State $\bu_t^* = \boldsymbol{\kappa}_t(\bx_t - \bx_t^k) + \bu_t^k$
            \State $\bx_{t+1} \sim p(*|\bx_t, \bu_t^*)$
        \EndFor
        \State Return $\bu^{k} = \{\bu_0^*, \bu_1^*, \cdots, \bu_{N_T}^*\}$
    \end{algorithmic}
\end{algorithm}

\subsection{Receding-Horizon Planning}
Given the multimodal CE trajectory optimization approach described above, we can now construct a receding-horizon planning algorithm, as described in Algorithm \ref{alg:nmpc}. 
\begin{algorithm}
\caption{Receding-Horizon Multimodal CE Planning}
\label{alg:nmpc}
\begin{algorithmic}[1]
    \State \textbf{Input:} $\bnu_0^*$
    \State $\bx_0 \gets$ \text{GetCurrentState}()
    \While{objective not completed}
        \State $\hat{\bnu}^* \gets$ \text{Optimize}$(\bnu_0^*, \bx_0)$ \Comment{See Alg.~\ref{alg:multi_modal}}
        \State $\bu^d \gets$ \text{MaximumLikelihoodEstimate}$(\bnu^*)$
        \For{$t = 0$ to $N_T$}
            \State $\bx^d_{t+1} = \bx^d_t + f(\bx^d_t, \bu^d_t) \Delta t$
        \EndFor
        \State $\bX^d \gets \{\bx^d_0, \bu^d_0, \bx^d_1, \bu^d_1, \ldots, \bu^d_{N_T}, \bx^d_{N_T+1}\}$
        \State \text{Execute}$(\bX^d)$
        \State $\bx_0 \gets$ \text{GetCurrentState}()
        \State $\bnu_0^* \gets$ \text{InitializePrior}$(\bnu^*, \bX^d, \bx_0)$ 
    \EndWhile
    \State \Return
\end{algorithmic}
\end{algorithm}

\section{Multi-robot Cross-Entropy Planning}
While planning multimodal policies can be particularly beneficial for navigating through complex static obstacle fields, it is also relevant for for multi-robot coordination. Here, we aim to leverage multimodal policies to achieve computationally tractable multi-robot planning by decoupling discrete high-level mode coordination from continuous low-level policy optimization. 

\subsection{Multimodal Policy Sharing}
For multi-robot planning, we assume communication among robot team members. At the end of each receding-horizon planning cycle, all robots synchronously share their current state and their multimodal policy distributions. In this way, a robot can generate predicted trajectories for its team members and evaluate inter-robot constraints during the next cycle of trajectory optimization. While this trajectory-sharing approach does not ensure joint synchronous optimization of robot control policies, it does reduce the computational requirements compared to a centralized policy optimization approach.

\subsection{Inter-robot Collision Constraints}
To estimate the probability of collision between robots, we check for collisions between a robot’s planned trajectories and the predicted trajectories of its neighbors, across all modes. Given the $i^{\textrm{th}}$ sampled trajectory from robot $j$, $\bX^{i}$, let $\bx^{n,m,k}_t$ be the state of the $n^{\textrm{th}}$ robot team member’s $m^{\textrm{th}}$ trajectory from the $k^{\textrm{th}}$ mode at time $t$, where $j\neq n$. We can estimate the probability of collision of trajectory $\bX^i$ as
\begin{align}
    g_A^{n, k}(\bX^i) &= \frac{1}{M} \sum_{m=1}^{M} g_A^{n, m, k}(\bX^i),
\end{align}
where
%We can follow the same indicator functions for each mode, and define
% \begin{align}
%     g_a^{n,m,k}(\bx_t^i) = \mathbb{I}\big( \mathrm{dist}(\bx_t^i, \bx_t^{n,m,k}) < L \big),\\
%     C_A^{n}(\bX^i) = \bigwedge_{k = 1}^KC_A^{n,k}(\bX^i)
% \end{align}
% where $C_A^{n,k}(\bX^i)$ is the constraint violation per mode. The trajectory only violates a constraint if it intersects with more than $P\%$ of trajectories in all modes independetly.

% To handle cases where other robots may follow multiple possible trajectory modes, we extend our collision avoidance constraint accordingly. Let $\bx_t^{n, m, k}$ denote the state of the  $n^{th}$  robot at time  $t$ on its $m^{th}$ trajectory sample from its $k^{th}$ policy mode.

% We define the pairwise collision indicator relative to each policy mode as
\begin{align}
    g_a^{n, m, k}(\bx_t^i) &= \mathbb{I}\Big( \mathrm{dist}\big(\bx_t^i, \bx_t^{n, m, k}\big) < L \Big),\\
    g_A^{n, m, k}(\bX^i) &= \bigvee_{t=0}^{N_T} g_a^{n, m, k}(\bx_t^i),
\end{align}
and $\bigvee$ is logical OR. We can then define a chance constraint as
\begin{align}
    C_A^{n, k}(\bX^i) &= \mathbb{I}\left( g_A^{n, k}(\bX^i) \geq P \right)
\end{align}
where probability $P\in(0,1)$.
A candidate trajectory $\bX^i$  is considered unsafe with respect to robot $n$ only if it violates the chance constraint in all modes of that robot, such that
\begin{align}
    C_A^n(\bX^i) = \bigwedge_{k=1}^K C_A^{n, k}(\bX^i),
\end{align}
where $\bigwedge$ denotes logical AND.

The overall multi-robot collision constraint is then
\begin{align}
    C_A(\bX^i) = \bigvee_{n \neq j} C_A^n(\bX^i).
\end{align}
Finally, the total constraint for $\bX^i$ is updated as
\begin{align}
    C(\bX^i) = c(\bx^i_0) \lor c(\bx^i_1) \lor \cdots \lor c(\bx^i_{N_T}) \lor C_A(\bX^i).
\end{align}

\subsection{Multimodal Policy Coordination}

Once each robot $i \in \{1, \cdots, N\}$ has optimized its candidate policies using the shared information from the \emph{prior} planning phase, a final coordination phase is conducted using the \emph{current} multimodal policies of all robots to maximize joint feasibility with respect to inter-robot collisions. 

Since individual robot policies are optimized using robot team information from the last planning phase, a robot cannot greedily select among its individual receding-horizon policies and ensure the chosen nominal trajectories will avoid inter-robot collisions. Therefore, at the conclusion of the current planning cycle, a centralized coordination algorithm finds the optimal set of policy modes.

Let $c = \{k_1, \cdots, k_N\}$ denote a selected set of policies, one for each robot, such that the nominal trajectory for robot $j$ is given by $\bX^j$. For each selection, we want to minimize the global cost while avoiding collisions:
\begin{align}
    c^* = \arg\min_{c} \sum_{j=1}^NJ(\bX^j)\\ 
    \textrm{s.t.}\quad g_a(\bx_t^j) >0 ;\;\forall\, t, j.
\end{align}
In the event that no candidate selection $c$ is feasible, we select $c^*$ to minimize the total number of constraint violations
\begin{align}
    c^* = \arg\min_{c} \sum_{t} \mathbb{I}(g_{a}(\bx^j_t) \le 0)
\end{align}
and break ties by using total cost $J_{\textrm{total}}=\sum_{i=1}^NJ(\bX^i)$. Here, we solve this via exhaustive search, but recognize that more computationally tractable approaches could be explored.
% Once each robot has independently optimized its candidate trajectories using the initial shared information, a final consensus phase is required to ensure collision-free and globally optimal execution. Individual robots selecting their locally optimal policy, do not guarantee a combination that avoids inter-robot collisions and minimizes the global cost jointly. To address this, all robots enter a consensus phase at the end of each planning cycle, sharing their complete set of candidate policies. With the independently optimized policies, a centralized controller jointly evaluates all possible combinations of policies, one for each robot, and identifies which set of policies satisfy the collision constraints. 
% \begin{align}
%     g_a(\bx_t^i) &= \{dist(\bx_t^i, \bx_t^{n,m}) - L < 0\} \ \forall \ n: i \neq n,%\\
% \end{align}
% The global cost for each combination is assessed, and the combination that is both constraint-free and minimizes the sum of local costs is selected for execution. Should all combinations violate collision constraints, the team executes the combination that results in the lowest cumulative number of constraint violations, thereby ensuring the least undesirable behavior under infeasibility, as shown in Algorithm \ref{alg:consensus}. 

\begin{algorithm}
\caption{Coordination of Joint Trajectory Set}
\label{alg:consensus}
\begin{algorithmic}[1]
    \State \textbf{Inputs:} $\{\mathcal{P}_i\}$: candidate policy sets for all robots
    % \State \textbf{Outputs:} Best joint policy combination, minimum pairwise safety distances
    \State $\mathcal{C} \gets$ Generate all possible combinations of policies %($K^{N_{\text{robots}}}$)
    \State Initialize best\_cost $\gets$ $\infty$; best\_combo $\gets$ None

    \For{each joint combination $\mathbf{c} \in \mathcal{C}$}
        \State total\_cost $\gets$ 0; total\_violations $\gets$ 0
        \For{each robot $i$ in $\mathbf{c}$}
            \State total\_cost $\mathrel{+}= J(\tau_i)$
            % \State total\_violations $\mathrel{+}= C_{\textrm{env}}(\tau_i)$ 
        \EndFor
        \For{all $(i, j)$ robot pairs}
            \For{each timestep $t$}
                \State Compute inter-robot distance $d_{ij}(t)$
                \If{$d_{ij}(t) < r_i + r_j$}
                    \State total\_violations $\mathrel{+}= 1$ 
                \EndIf
            \EndFor
        \EndFor
        \State cost\_with\_penalty $\gets$ total\_cost + $\lambda \cdot$ total\_violations
        \If{cost\_with\_penalty $<$ best\_cost}
            \State best\_cost $\gets$ cost\_with\_penalty
            \State best\_combo $\gets$ $\mathbf{c}$
        \EndIf
    \EndFor
    
    \State \textbf{Return:} best\_combo
\end{algorithmic}
\end{algorithm}
\vspace{-15pt}

%% file: sim_experiments.tex
\section{Simulation Experiments}

\begin{figure*}[t]
\vspace{5pt}
\centering
  \includegraphics[width=\textwidth]{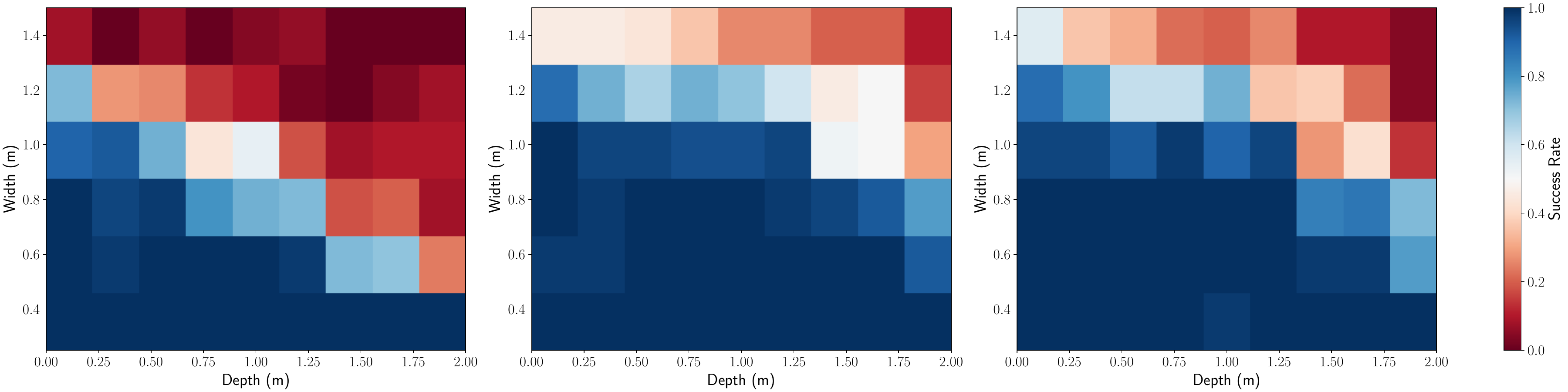}
  \caption{Success rate heatmaps across different trap parameters. From left to right: single-mode CE, two-mode CE, and two-mode CE without TVLQR warm-starts. Success is defined as reaching the goal within 10 seconds without colliding with obstacles. The two-mode CE approach outperformed the single-mode CE in environments with moderately deep and wide traps. Furthermore, incorporating TVLQR-based warm starts for the secondary modes led to additional performance gains in the deepest and widest trap configurations.}
  \label{fig:Heatmap}
  \vspace{-5pt}
\end{figure*}

We conducted extensive Monte Carlo simulations to evaluate the effectiveness of our proposed multimodal cross-entropy planner in two scenarios: navigation in trap environments with many local minima and distributed multi-robot collision avoidance.
\subsection{Dynamics Model}
Each agent is modeled as a stochastic bicycle with acceleration and steering rate inputs. The agent's state vector is defined as ${\bx_t} = [p_x, p_y, \theta, v, \delta_s]^T$, comprising position, heading, velocity, and steering angle. The control vector is ${\bu_t} = [\dot{v}, \dot{\delta_s}]^T$, with wheelbase $l = 0.33\,\mathrm{m}$. Robot dynamics are given by $\bx_{t+1} \sim p(\cdot | \bx_t, \bu_t) = \bx_t + [f(\bx_t, \bu_t) + \vectg{\omega}] \Delta t$ where
$f(\bx_t, \bu_t) = [v\cos(\theta),\ v\sin(\theta),\ v\tan(\delta_s)/l,\ \dot{v},\ \dot{\delta}_s]^T$, $\vectg{\omega} \sim \mathcal{N}(\mathbf{0}, \boldsymbol{\Gamma})$, and $\boldsymbol{\Gamma} = \mathrm{diag}([0.001,\ 0.001,\ 0.012,\ 0.1,\ 0.006])$. Here $\vectg{\omega}$ is Gaussian noise and $\boldsymbol{\Gamma}$ is an estimation of the process covariance. The acceleration is limited to $\dot{v} \in [-1,1]\ \mathrm{m/s^2} $, the steering rate is limited to $\dot{\delta_s} \in [-1,1]\ \mathrm{rad/s}$, the velocity is limited to $v \in [-0.5,2]\ \mathrm{m/s} $, and the steering angle is limited to $\delta_s \in [-0.4,0.4]\ \mathrm{radians}$. 

The planner and environment hyperparameters are summarized in Table~\ref{tab:params}. 
\begin{table}[ht]
\caption{Planner and Cost Function Parameters}
\centering
\begin{tabular}{||l | l||}
\hline
\textbf{Parameter} & \textbf{Value} \\
\hline
Number of samples ($M$)            & $1024$ \\
Elite fraction ($\rho\%$)          & $10\%$ \\
Planning horizon ($T$)             & $40$ time steps (2 seconds)\\
Time step ($\Delta t$)             & $0.05~\mathrm{s}$ \\
Stage cost weight ($Q$)            & $\mathrm{diag}([1,\, 1,\, 0,\, 0,\, 0])$ \\
Terminal cost weight ($Q_f$)       & $\mathrm{diag}([40,\, 40,\, 0,\, 0,\, 0])$ \\
Control input cost ($R$)           & $\mathrm{diag}([0.1,\, 0.1])$ \\
\hline
\textbf{TVLQR Weights} & \\
Stage state cost ($Q$) & $\mathrm{diag}([10,\, 10,\, 1,\, 1,\, 1])$ \\
Control input cost ($R$) & $\mathrm{diag}([1,\, 1])$ \\
Terminal state cost ($Q_{f}$) & $\mathrm{diag}([100,\, 100,\, 10,\, 10,\, 10])$\\
% Terminal state cost ($Q_{f}$) & 10 * $Q$\\
\hline
\end{tabular}
\label{tab:params}
\end{table}
\vspace{-13pt}

\subsection{Trap Environment}
To evaluate planner robustness against local minima, we conducted a total of 2,700 ``trap environment'' trials. These trials spanned 54 distinct U-shaped obstacle configurations, systematically varying trap width and depth. For each obstacle type, we ran 50 randomized trials with unique initial and goal state pairs. In every trial, initial and goal positions were randomly sampled within the workspace bounds, $(p_x, p_y) \in [(-1,-6),\,(11,6)]$ meters, with resampling to enforce a minimum Euclidean distance of $7$ meters between them. The agent’s initial heading was oriented toward the goal, while its initial velocity and steering angle were sampled uniformly from $v \in [-0.5,\,2.0]\,\mathrm{m/s}$ and $\delta_s \in [-0.1,\,0.1]\,\mathrm{radians}$, respectively.

Each trap environment contained a single U-shaped obstacle, created by overlapping circles of radius $0.25$ meters and positioned two-thirds of the distance from the start to the goal configurations, with the cavity facing the initial robot position. Throughout our study, trap widths and depths were varied in increments of $0.25$ meters, spanning $0.25$ to $1.5$ meters for width and $0.0$ to $2.0$ meters for depth, resulting in six discrete widths, nine discrete depths, and a total of 54 distinct trap geometries for evaluation. Success was defined as the agent reaching the goal within $10$ seconds without colliding with the trap obstacle.

As shown in Figure~\ref{fig:Heatmap}, the two-mode planner outperformed the one-mode planner in avoiding deeper and wider traps. The two-mode planner also performed similarly regardless of whether TVLQR warm-starts were used in thin and shallow traps. However, TVLQR warm-starts provided a distinct advantage when navigating deeper and wider traps.
% \begin{figure}
%     \centering
%     \includegraphics[width=0.85\columnwidth]{images/Mode3_SuccessHeatmapv2.pdf}
%     \caption{Heatmap of success rate for two modes but without warm-starting the secondary modes with TVLQR. Success was defined as the agent reaching its goal within 10 seconds, without colliding with obstacles.}
%     \label{fig:Trap2ModeNoTVLQR}
% \end{figure}
% \vspace{-25pt}
\begin{figure}
    \centering
    \includegraphics[width=\columnwidth]{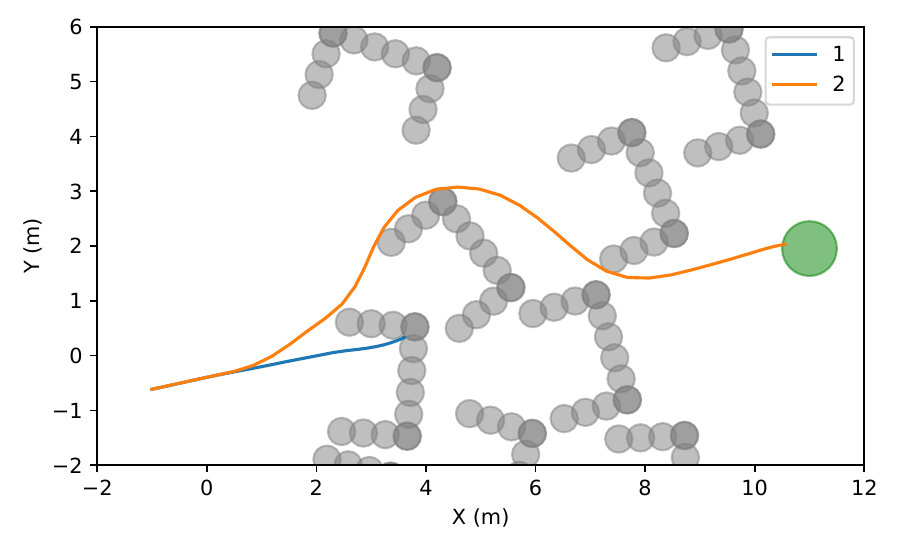}
    \caption{In the trap environment, single-mode cross-entropy optimization fails due to local minima, but employing a two-mode policy successfully navigates to the goal.}
    \label{fig:TrapEnvionment}
    \vspace{-18pt}
\end{figure}

% To further compare the performance of single-mode and multimodal policies, we generated a diverse set of trap environments. For each environment, the agent's initial position was uniformly sampled from $[(-1,-6),(0,6)]$, and the goal position from $[(10,-6),(11,6)]$. The initial heading was set to point directly toward the goal, with zero velocity and steering angle. 12 U-shaped obstacles were randomly placed throughout the workspace, with their centers uniformly sampled from $[(1,-6),(10,6)]$. Each U-shaped obstacle—consisting of circles with radius $0.25\,\mathrm{m}$—was oriented with its cavity directed toward the initial position with an orientation variance of $\pm \pi/8\,\textrm{radians}$. Of the $650$ generated environments, $308$ were rejected due to no feasible path between start and goal. For each of the remaining $342$ environments, eight trials were run: agents generated $1024$ and $2048$ samples and clustered them into $K$ modes, for $K=\{1,2,3,4\}$. Figure \ref{fig:TrapEnvionment} shows a path where two modes succeeded and 1 mode got caught in a local minima. As shown in Table~\ref{tab:modes_performance}, increasing the number of modes led to improved performance up to three modes, but performance declined with four modes in this environment. This decrease may be attributed to a reduction in effective exploration, as the fixed number of samples becomes spread too thinly across an increasing number of modes.

To further compare single-mode and multimodal policies, we generated a diverse set of trap environments. In each, the agent's start was uniformly sampled from $[(-1,-6),(0,6)]$ meters and the goal from $[(10,-6),(11,6)]$ meters, with heading set toward the goal and zero initial velocity/steering. Twelve U-shaped obstacles (composed of $0.25$ meter-radius circles) were randomly placed with their cavities facing the start, each oriented within $\pm \pi/8$ radians. Out of $650$ generated environments, $342$ with feasible solutions were retained. For each environment, six trials were conducted: the robot generated $1024$ samples for optimizing $K = 1$ and $K = 2$ modes, and $2048$ samples for optimizing $K = 1$, $2$, $3$, and $4$ modes. Figure~\ref{fig:TrapEnvionment} illustrates that multiple modes can help avoid local minima where single-mode policies get trapped. As shown in Table~\ref{tab:modes_performance}, performance improved as $K$ increased up to three modes, but declined with four modes—likely because samples became too thinly distributed across clusters, reducing effective exploration.

% \begin{table}[ht]
% \centering
% \caption{Performance Across Different Numbers of Modes}
% \begin{tabular}{||c|cccc|}
% \hline
% \# Samples & 1 Mode & 2 Modes & 3 Modes & 4 Modes \\
% \hline\hline
% 1024     & 122    & 251     & X       & X       \\
% \hline
% 2048     & 181    & 258     & 269     & 255     \\
% \hline
% \end{tabular}
% \label{tab:modes_performance}
% \end{table}

\begin{table}[ht]
\centering
\caption{Performance Across Different Numbers of Modes}
\begin{tabular}{||c|cccc|}
\hline
\# Samples & 1 Mode & 2 Modes & 3 Modes & 4 Modes \\
\hline\hline
1024     & 35.7\%    & 73.4\%     & X       & X       \\
\hline
2048     & 52.9\%    & 75.4\%     & 78.7\%     & 74.7\%     \\
\hline
\end{tabular}
\label{tab:modes_performance}
\end{table}
% \vspace{-10pt}

% \begin{table}
%   \centering
%   % \vspace{8pt}
%   \begin{tabular}{||c| c c c||}
%     \hline
%     \textbf{\thead{Trial Type}}  & \textbf{\thead{Number of\\ Collisions}} & \textbf{\thead{ Leader \\Constraint }} & \textbf{\thead{ Follower \\Constraint}}\\ [0.5ex]
%     \hline\hline
%     % \thead{PACNMPC\\50} & 1 & 0.00856 & 0.01288\\
%     % \hline
%     \thead{PAC-NMPC} & 0 & 0.00465 & 0.013 \\
%     \hline
%     \thead{RA-MPPI} & 38 & 0.0521 & 0.1788\\ [1ex]
%     \hline
%     \thead{PAC-NMPC State \\Noise Blind} & 26 & 0.1208 & 0.148\\
%     \hline
%     \thead{PAC-NMPC State \\Noise Aware} & 2 & 0.0987 & 0.1854\\
%     \hline
%     \thead{RA-MPPI State \\Noise Aware} & 33 & 0.208 & 0.289\\ [1ex]
%     \hline
%     %\thead{Mean Final State\\200} & 3 & 0.00669 & 0.0148 \\
%     %\hline
%    % \thead{Follower RRT\\ RRT 200} & 8 & 0.00296 & 0.0487 \\
%     %\hline
%   \end{tabular}
%   \caption{Collisions and Average Probability of Constraint Violations}
%   \label{tab:NT}
%   \vspace{-15pt}
% \end{table}

% To further show the difference in performance between a single mode and a bimodal policy, we made 650 Trap environments \textcolor{red}{We sampled from blah and blah}. We rejected 308 trials dues to no path being able to be found. On the 342 remaining trials 1 modes completed 122 (35.7\%) and 2 modes completed 251 (73.4\%).

\subsection{Multi-Agent Collision Avoidance Trials}
\begin{figure}
    \centering
    \vspace{5pt}
    \includegraphics[width=\columnwidth]{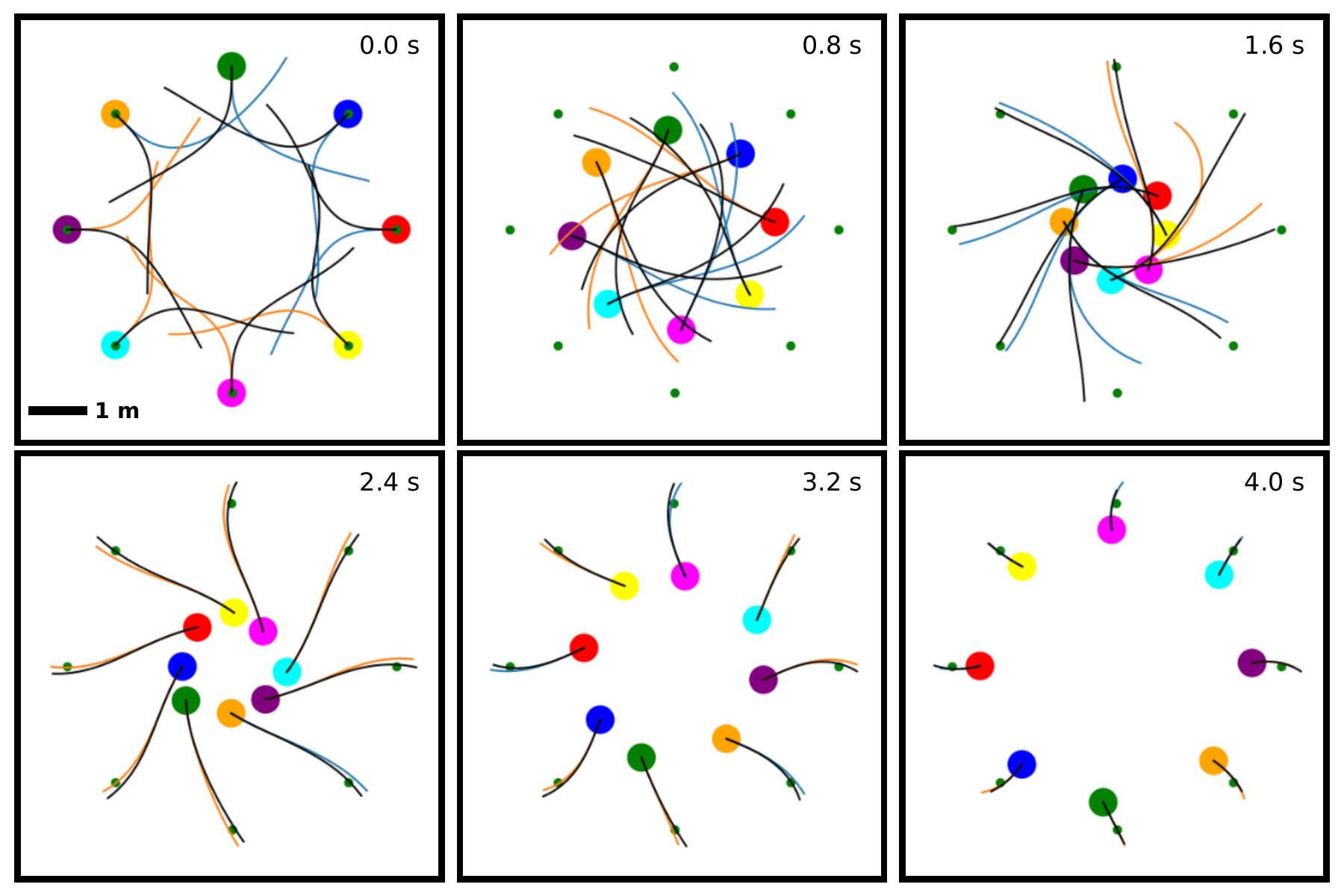}
    \caption{Trajectories of eight robots in a 3-meter radius antipodal simulation, optimizing a two-mode policy. Blue and orange denote the order in which candidate policies were provided to the centralized controller; black indicates the final policy selected for execution.}
    \label{fig:8AgentGif}
\end{figure}

\begin{figure}
    \centering
    \includegraphics[width=\columnwidth]{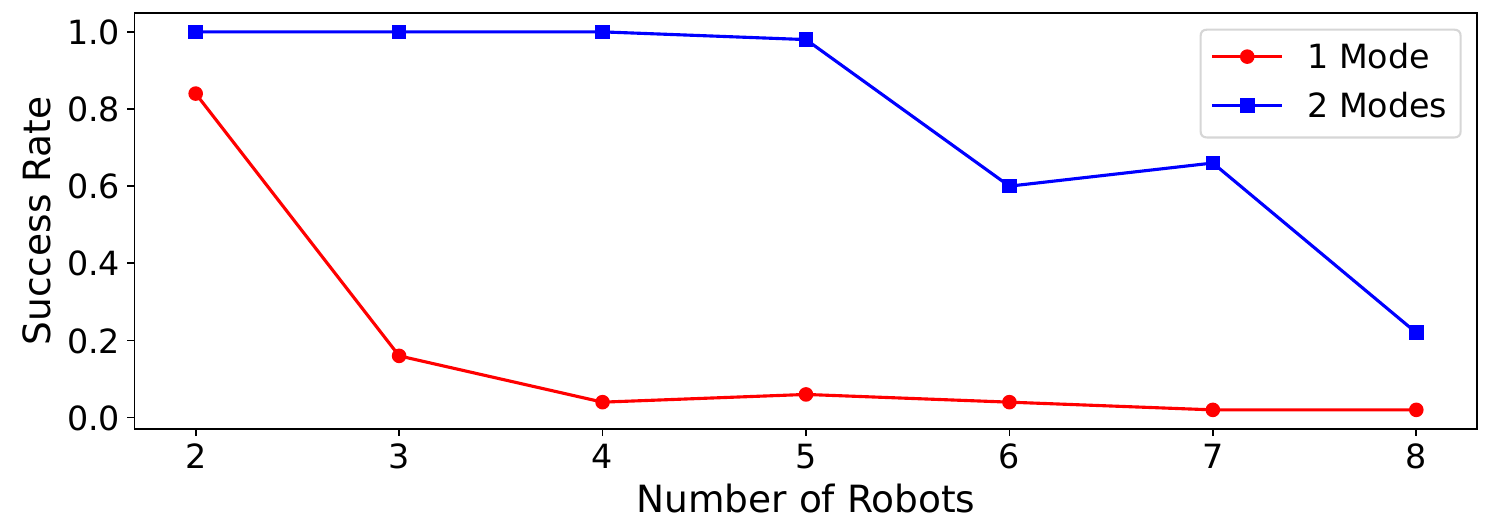}
    \caption{Results from multi-robot antipodal experiments, where robots were initialized uniformly on a circle of radius three meters. The two-mode case achieved perfect performance with four agents and also had a higher success rate than the single-mode case for all numbers of robots.}
    \label{fig:AntipodalData}
\end{figure}

Inter-agent collision experiments were conducted for sets of two to eight agents, with 50 trials each for each set. Agents were initialized at equal intervals along the circumference of a circle with a radius of $3$ meters, each oriented radially inward. Their goals were positioned at antipodal points on the circle. This scenario tested the system's robustness in identifying distinct modes and employing policy coordination to select a collision-free trajectory set as the difficulty increased with the addition of more agents. An example with eight agents in a two-mode trial is shown in Figure \ref{fig:8AgentGif}. As shown in Figure~\ref{fig:AntipodalData}, the single-mode approach experiences a sharp decline in the number of collision-free paths found as the number of agents increases. In contrast, the two-mode approach maintains a high success rate up to five agents before its performance begins to drop. Notably, even with eight agents, the two-mode approach outperforms the single-mode case with only three agents.

% \begin{figure}
%     \centering
%     \includegraphics[width=\columnwidth]{images/AntipodalModesv2.eps}
%     \caption{Results from this Experiments, showing 1 vs 2 modes on the different traps? (Maybe Difference between 1 mode and 2 and 3 modes}
%     \label{fig:placeholder}
% \end{figure}

% \subsection{Parameters}

%% file: hardware_experiments.tex
\section{Hardware Experiments}
To validate our approach in real-world conditions, we deployed the multimodal CE planner on a physical robotic platform. We used a $1/10^\textrm{th}$ scale Traxxas Rally Car whose dynamics were modeled as a stochastic bicycle system with a wheelbase of $l = 0.354$ meters. The velocity was constrained to $v \in [-0.5,1.0]~\mathrm{m/s}$, and the steering angle limited to $\delta_s \in [-0.3,0.3]$ radians.

% We obtained state estimates using a motion capture system, generated control commands based on the planned trajectories, and executed the commands via TVLQR, all in real-time. We tested the approach in a single-obstacle trap environment, and added two virtual agents to evaluate a three-agent antipodal navigation scenario. All planning and policy computations were run on an Nvidia 4070 GPU at 5 Hz using distinct computers. The on-board Jetson Orin AGX was used to follow the planned trajectory using TVLQR. A photo of the car can be seen in Figure \ref{fig:CarImage}.

\begin{figure}
    \centering
    \vspace{5pt}
    \includegraphics[width=0.7\columnwidth]{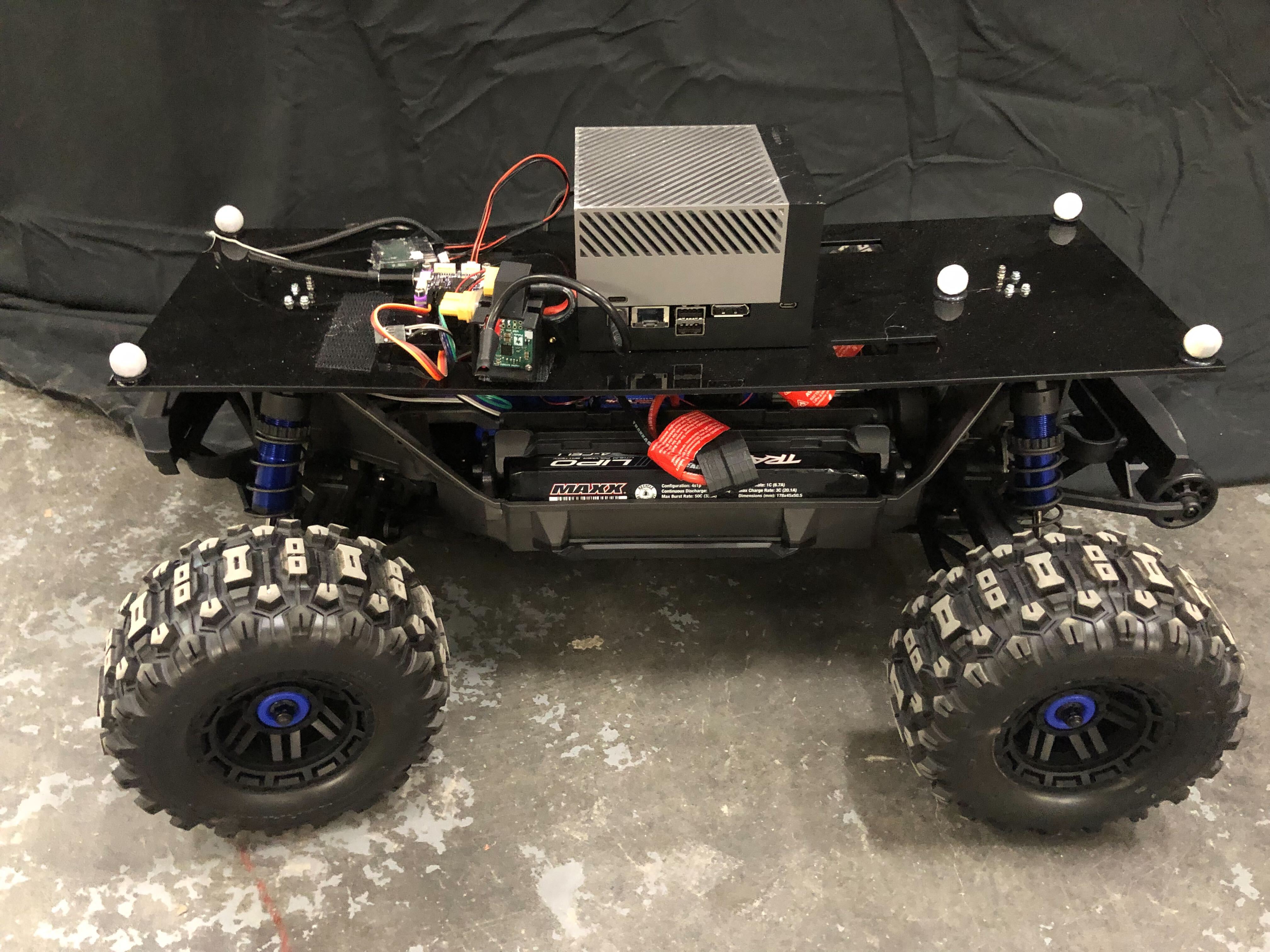}
    \caption{Photo of $1/10^\textrm{th}$ scale Traxxas Rally Car}
    \label{fig:CarImage}
\end{figure}

State estimates were obtained via motion capture, with control commands generated from planned trajectories and executed in real-time using TVLQR. We validated our approach in both a single-obstacle trap environment and a three-agent antipodal navigation scenario with two virtual agents. Planning and policy computations were performed at 5~Hz on separate computers with an Nvidia 4070 GPU, while a Jetson Orin AGX on board the vehicle executed the TVLQR controller. A photo of the car is shown in Figure~\ref{fig:CarImage}.

% \subsection{Dynamics Model}
% Each agent is modeled as a stochastic bicycle, with acceleration and steering rate inputs. The agent state vector is defined as ${\bx_t} = [p_x, p_y, \theta, v, \delta_s]^T$, comprising position, heading, velocity, and steering angle. The control vector is ${\bu_t} = [\dot{v}, \dot{\delta_s}]^T$, with $l = 0.33\,\mathrm{m}$ as the wheelbase. Agent dynamics are given by:
% \begin{align}
%   \bx_{t+1} &\sim p(\cdot | \bx_t, \bu_t) = \bx_t + [f(\bx_t, \bu_t) + \vectg{\omega}] \Delta t \\
%   f(\bx_t, \bu_t) &= [v\cos(\theta),\ v\sin(\theta),\ v\tan(\delta)/l,\ \dot{v},\ \dot{\delta}]^T \\
%   \vectg{\omega} &\sim \mathcal{N}(\mathbf{0}, \boldsymbol{\Gamma}) \\\quad
%   \boldsymbol{\Gamma} &= \mathrm{diag}([0.001,\ 0.001,\ 0.1,\ 0.2,\ 0.001])
% \end{align}
% The acceleration input is limited to $-1\ m/s^2 \leq \dot{v} \leq 1\ m/s^2 $, the steering rate input is limited $-1\ rad/s \leq \dot{\delta_s} \leq 1\ rad/s$, the velocity is limited to $-0.5\ m/s  \leq v \leq 2\ m/s $, the steering angle is limited to $-0.4\ rad \leq \delta_s \leq 0.4\ rad$. 
\begin{figure}
    \centering
    \includegraphics[width=\columnwidth]{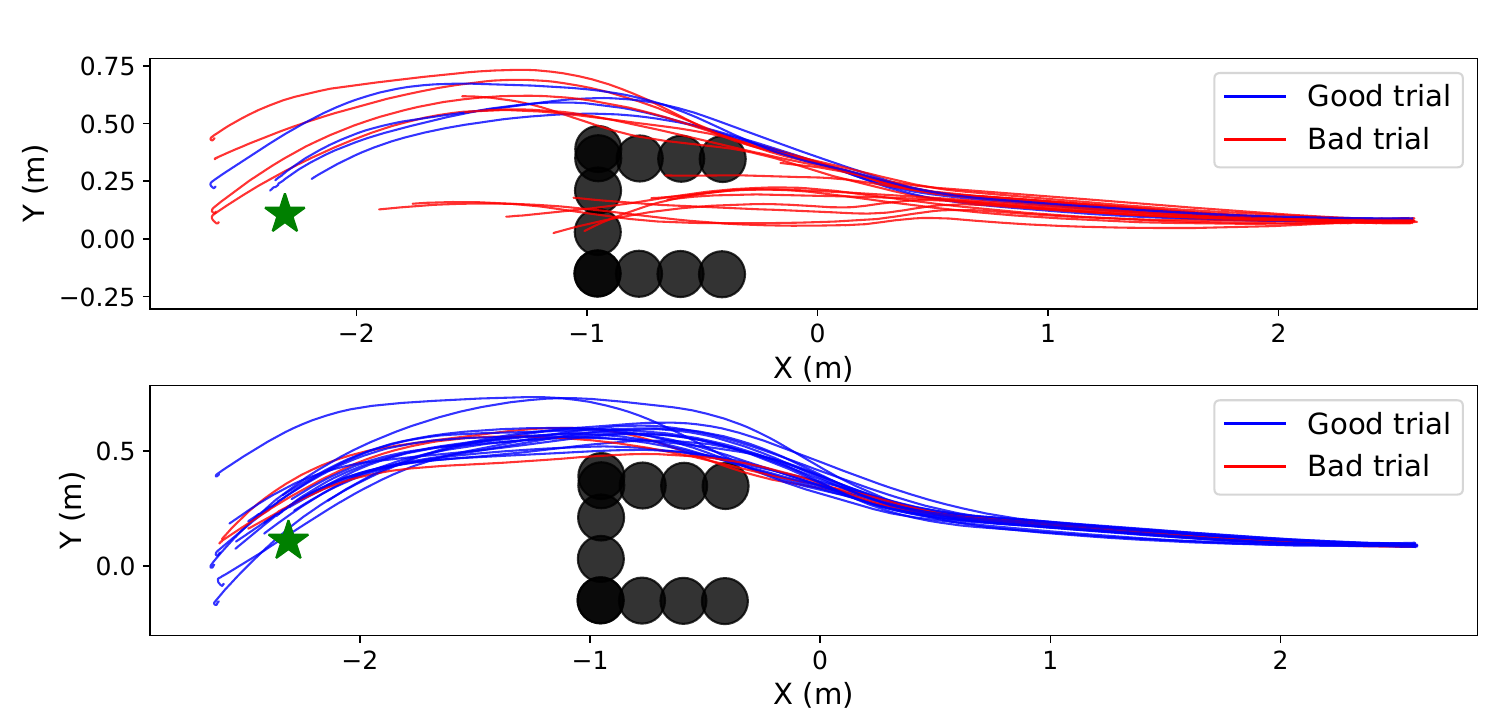}
    \caption{Trap Environment Hardware Trials, (Top) 1 Mode: 3 Trials made it to the goal collision-free as seen in blue, and nine trials got caught in the local minima. (Bottom) 2 Modes:  17 Trials made it to the goal collision-free as seen in blue, and all tries avoided the local minima.}
    \label{fig:HardwareSingleAgent}
    \vspace{-5pt}
\end{figure}

We tested the hardware using a virtual trap environment with a width of 0.5 meters and a depth of 0.6 meters. Figure \ref{fig:HardwareSingleAgent} shows the trajectories from the single mode experiment, where only 20\% of the trials made it to the goal while being collision-free. We can see that nine trajectories become trapped in the local minima, whereas in the two-mode experiment, no trajectories are caught in the trap, and they succeed with an 85\% success rate.
% \begin{figure}
%     \centering
%     \includegraphics[width=\columnwidth]{images/2ModeSuccess.pdf}
%     \caption{2 Modes Hardware Trials, 17 Trials made it to the goal collision free as seen in blue, and all tries avoided the local minima.}
%     \label{fig:2ModesHardware}
% \end{figure}
We further evaluated the antipodal scenario by testing a real robot in conjunction with two virtual robots on a circle with a radius of $1.7$ meters, keeping all other parameters constant. In this setting, the two-mode approach achieved collision-free trajectories in eight out of ten trials, whereas the single-mode approach succeeded in only one trial. Table \ref{tab:hardware_performance} shows the performance of the hardware trials.

\begin{table}[H]
\centering
\vspace{5pt}
\caption{Antipodal Hardware Performance}
\begin{tabular}{||c|ccc||}
\hline
\makecell{Number \\ of Modes} & Success Rate & \makecell{Average Minimum \\ Distance (m)} & \makecell{Median Minimum \\ Distance (m)} \\
\hline\hline
1     & 10\%   & 0.184     & 0.136\\
\hline
2     & 80\%    & 0.474     & 0.504\\
\hline
\end{tabular}
\label{tab:hardware_performance}
\end{table}
\vspace{-10pt}

%% file: discussion.tex
\section{Discussion}

In this paper, we introduced a distributed receding-horizon stochastic planning framework capable of navigating agents through trap environments and resolving multi-agent trajectory conflicts. By applying K-means clustering to sampled trajectories, we identified up to four distinct policies that enabled agents to escape local minima. We further demonstrated that TVLQR can be utilized to warm-start secondary policies, thereby preserving policy diversity and enhancing optimization efficiency. Our multi-modal trajectory optimization approach enabled robust deconfliction of trajectories among multiple agents, demonstrating high success rates and real-time performance on hardware. Future work includes developing improved mode identification and selection strategies, as well as extending the framework to incorporate sensor fusion and planning under environmental uncertainty.

% \section*{ACKNOWLEDGMENT}
% We gratefully acknowledge the support of the Army Research Laboratory under grant W911NF-22-2-0241. 